\def\year{2022}\relax
\documentclass[letterpaper]{article} 
\usepackage{aaai22}  
\usepackage{times}  
\usepackage{helvet}  
\usepackage{courier}  
\usepackage[hyphens]{url}  
\usepackage{graphicx} 
\usepackage{natbib}  
\usepackage{caption} 
\DeclareCaptionStyle{ruled}{labelfont=normalfont,labelsep=colon,strut=off} 
\usepackage{times}
\usepackage{latexsym}

\usepackage{graphicx}
\usepackage{microtype}
\usepackage{adjustbox}
\usepackage{booktabs}
\usepackage{makecell}
\usepackage{amsfonts}
\usepackage{amsmath}
\allowdisplaybreaks[4]
\usepackage{url}
\usepackage{multicol}
\usepackage{multirow}
\usepackage{graphicx}
\usepackage{algorithm}
\usepackage{algorithmicx}
\usepackage{algpseudocode}
\usepackage{tabularx}
\usepackage{newfloat}
\usepackage{listings}
\floatstyle{ruled}
\newfloat{listing}{tb}{lst}{}
\floatname{listing}{Listing}
\title{Improving Logical-Level Natural Language Generation with Topic-Conditioned Data Augmentation and Logical Form Generation}

\author {
   Ao Liu,\textsuperscript{\rm 1}
   Congjian Luo, \textsuperscript{\rm 2}
   Naoaki Okazaki     \textsuperscript{\rm 1}
}
\affiliations {
    \textsuperscript{\rm 1} Tokyo Institute of Technology\\
    \textsuperscript{\rm 2} University of Electronic Science and Technology of China\\
    liu.ao@nlp.c.titech.ac.jp, 
     im.congjian@gmail.com,
    okazaki@c.titech.ac.jp
}

\usepackage{bibentry}

\begin{document}

\maketitle

\begin{abstract}
    Logical Natural Language Generation, i.e., generating textual descriptions that can be \textit{logically entailed} by a structured table, has been a challenge due to the low fidelity of the generation. \citet{chen2020logic2text} have addressed this problem by annotating interim logical programs to control the generation contents and semantics, and presented the task of \textbf{table-aware} logical form to text (Logic2text) generation. However, although table instances are abundant in the real world, logical forms paired with textual descriptions require costly human annotation work, which limits the  performance of neural models. To mitigate this,  we propose topic-conditioned data augmentation (TopicDA), which utilizes GPT-2 to generate unpaired logical forms and textual descriptions directly from tables. We further introduce logical form generation (LG), a dual task of Logic2text that requires generating a valid logical form based on a text description of a table. We also propose a semi-supervised learning approach to jointly train a Logic2text and an LG model with both labeled and augmented data. The two models benefit from each other by providing extra supervision signals through back-translation. Experimental results on the Logic2text dataset and the LG task demonstrate that our approach can effectively utilize the augmented data and outperform supervised baselines by a substantial margin.
\end{abstract}

\section{Introduction}
\label{sec:intro}
Natural language generation (NLG) from structured data has been a long-standing research problem. Traditional NLG datasets~\cite{novikova2017e2e,lebret2016wikibio} focused on surface-level realization of superficial facts in the structured data. Recently, \citet{chen2020logicalNLG} released the LogicNLG dataset, which requires the generation of textual descriptions that can be \textit{logically entailed} by a table. However, deep models built on this dataset exhibited the problems of low fidelity and uncontrollable content selection. \citet{deconfounded} proposed variational models to enhance the logical fidelity of the generated sentences, but still presented low fidelity scores on human evaluation.

\begin{figure}[t]
    \centering
    \includegraphics[width=\linewidth]{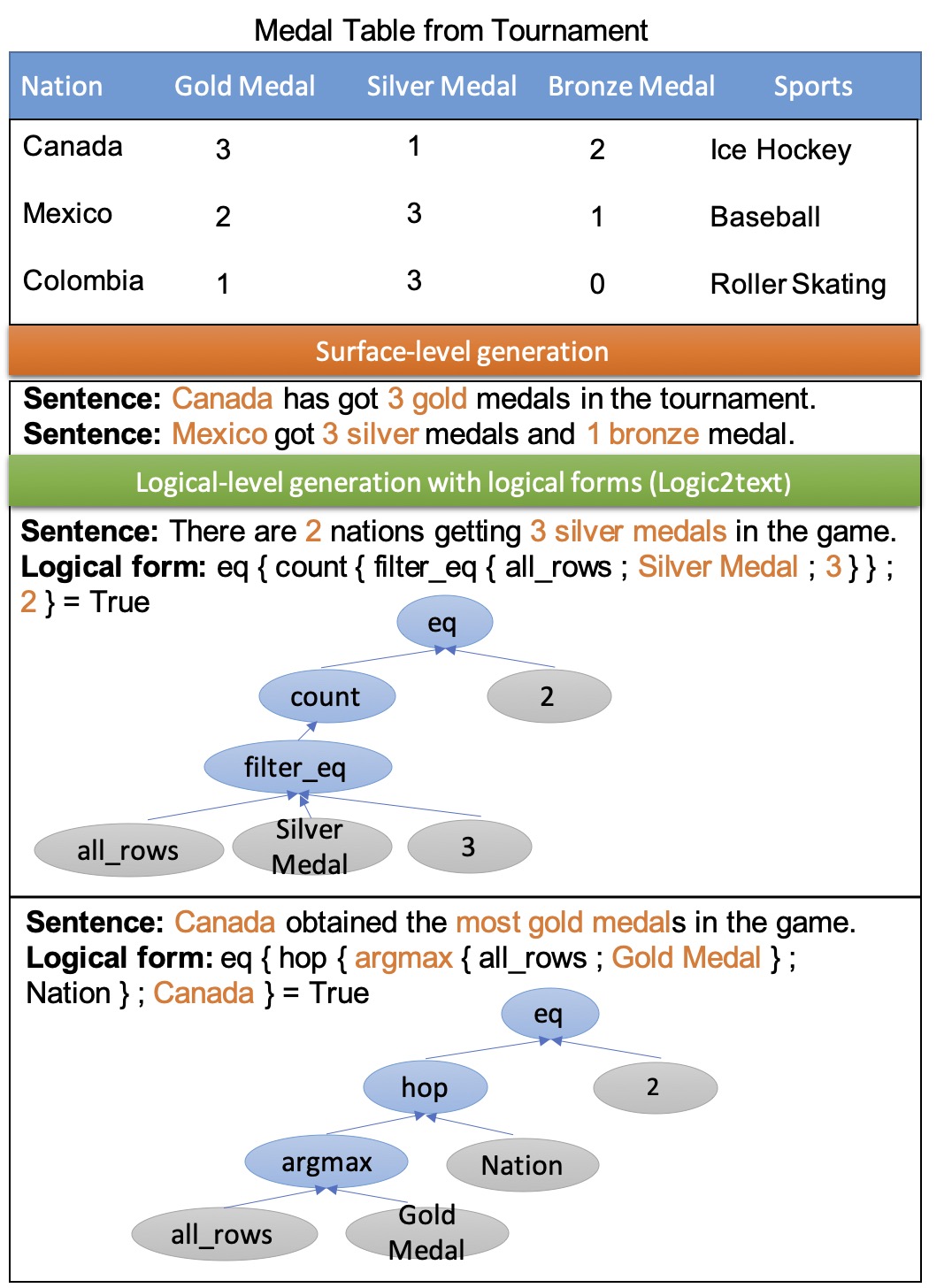}
    \caption{Comparison between surface-level NLG and logical-level NLG (Logic2text). Logic2text generates logical descriptions from a table based on an annotated logical form with diverse logic types. The two examples here are with the logic type \textit{count} and \textit{superlative}. Function nodes are shown in blue. This example was modified from ~\citet{chen2020logicalNLG}.}
    \label{fig:example}
\end{figure}

An effective remedy for this is to annotate high-quality mediators to guide the generation. \citet{chen2020logic2text} proposed the logical-level NLG task and released another dataset referred to as Logic2text. This task requires generating a sentence based on both a table and a logical form, which promotes faithful and controllable generation compared with LogicNLG~\cite{chen2020logicalNLG}. They annotated logical forms paired with corresponding textual descriptions, which resulted in an exciting boost in terms of the human-evaluated fidelity score from 20.2\% to 80.4\%. An example of the logical-level NLG (Logic2text) is depicted in Figure \ref{fig:example}, along with a comparison to the surface-level NLG.

\begin{figure*}[t]
    \centering
    \includegraphics[width=\linewidth]{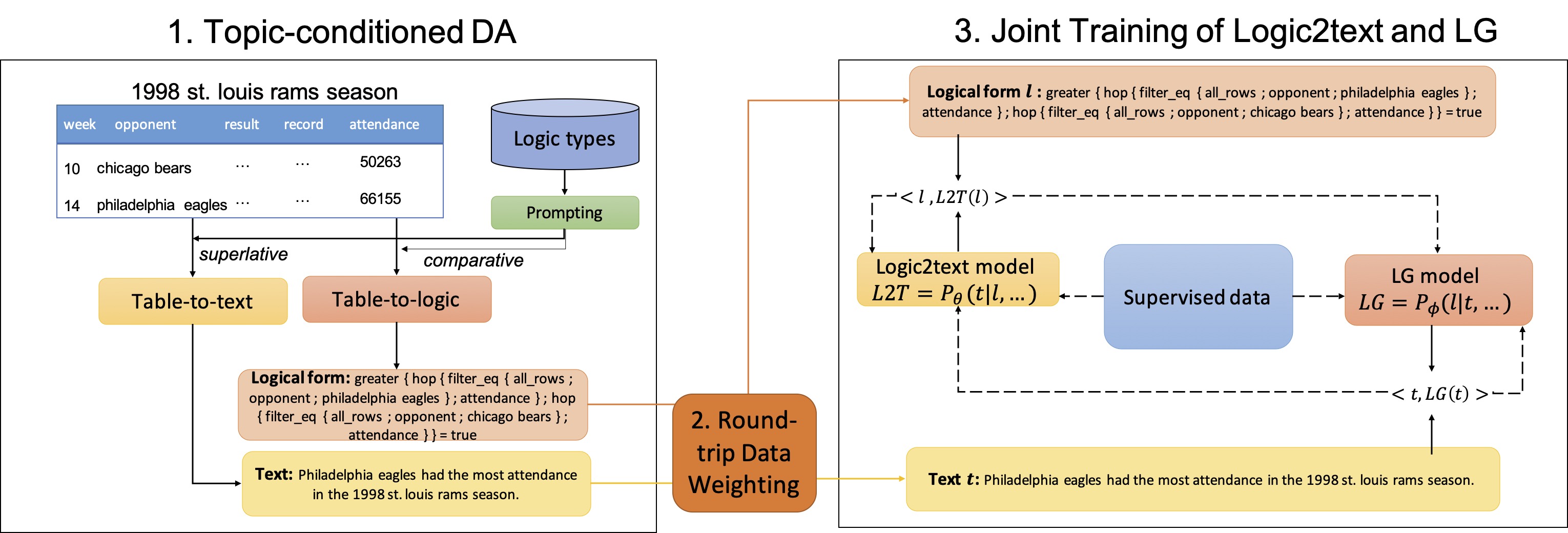}
    \caption{An overview of the proposed frame work, including (1) Topic-conditioned data augmentation (TopicDA), (2) Round-trip data weighting and (3) joint training of Logic2text and LG models. The logical forms and texts augmented in (1) are weighted in (2) and fed to (3) as unpaired data for pseudo-parallel data construction.
    In (3), the solid arrows indicate that the L2T/LG models generate pseudo-parallel data with the augmented monolingual logical-forms/texts. The dashed arrows indicate that the generated pseudo instances are used to train the models via back-translation and self-training. Note that the table in stage (1) is the generation context for L2T/LG generation in stage (3), but we ignore it here for simplicity of illustration. The illustration of (3) is inspired by \cite{guo2020revisiting}.}.
    \label{fig:framework}
\end{figure*}

Nevertheless, the labor-intensive human work of pairing logical forms and textual descriptions limited the scale of the Logic2text dataset (ca. 10.8k instances), which is much smaller than those of the common benchmarks on the surface-level NLG~\cite{novikova2017e2e,lebret2016wikibio}.
Pre-trained models such as GPT-2~\cite{radford2019gpt2} could work on the small amount of supervision data, using the rich contextual knowledge learnt from large-scale corpora; however, there is a lot of room for improvement in terms of the generation quality~\cite{chen2020logic2text}.
Moreover, we observe that each table in the Logic2text dataset is only associated with at most 3 examples. However, a table can contain abundant logical-level facts derived by various logical operations, while the dataset only covers a limited part of them. 
Inspired by this, we propose \textit{topic-conditioned data augmentation} (TopicDA) to bootstrap synthetic examples (i.e., unpaired logical forms (LFs) and texts) from existing supervised data. Specifically, we train auxiliary topic-conditioned \textit{table-to-logic} and \textit{table-to-text} models by fine-tuning GPT-2 on the supervised data to generate additional logical forms and texts directly from tables. By providing the models with pre-defined logic types as \textit{topics}, we generate LFs and texts with diverse logic types even from tables appearing in the original training data. As depicted in Figure \ref{fig:framework}, when we assign different topics such as \textit{superlative} and \textit{comparative} to an input table, the DA models can generate new logical forms or texts consistent with the given topics. Finally, we are able to mine more logical-level facts from existing tables for data augmentation without resorting to any additional resource.

Additionally, we introduce \textit{logical form generation (LG)}, a dual task of Logic2text that requires generating a valid logical form based on a text description and a corresponding table. Inspired by previous works on the joint learning of dual NLP tasks~\cite{chang2021neural, chang2021jointly, qader2019semi, guo2020cyclegt, schmitt2020unsupervised}, we propose the simultaneous solution of the Logic2text task and LG task by iteratively generating pseudo parallel data from the augmented data.

A subsequent challenge is that some of the augmented data can be noisy and impair the performance of the semi-supervised learning. We thus incorporate a \textit{round-trip data weighting} strategy to balance the weights of the different unpaired samples. We employ the round-trip BERTScore~\cite{bert-score} to evaluate the qualities of the augmented data and weight them during the joint training. We also adopt \textit{curriculum learning}~\cite{bengio2009curriculum} to further improve the joint training.

We evaluate the proposed methods on the Logic2text dataset and its dual task LG, conducting experiments under two different settings: (1) Full data: we exploit all the supervised data for both training DA models and joint training, (2) Few-shot: we randomly sample only 1000 instances for DA and joint training. Experimental results on both automatic and human evaluation demonstrate the effectiveness of the proposed framework in leveraging augmented data. 
Furthermore, analysis experiments on data augmentation demonstrate that the proposed TopicDA method can generate topic-diversified data with reasonable validity. Additionally, we find that the LG model can produce silver logical form annotations for the LogicNLG benchmark~\cite{chen2020logicalNLG}, suggesting that the proposed LG task can promote future work on the development of new logical-level NLG benchmarks.

\section{Task Formalization}
\label{sec:task}
In \textbf{Logic2text}, an input consists of a table $d$ and a logical form $l$ that can be executed on $d$, and an output is a sentence description $t=[w_1, w_2, \dots, w_n]$. We aim to train a model $P_{\theta}(t|l, d)$ to generate $t^*$ that can be supported by both the table $d$ and logical from $l$. For \textbf{Logical Form Generation (LG)}, we train a model $P_{\phi}(l|t, d)$ to estimate $l^*$ from the input description $t$, which is also supported by the table $d$. LG is the inverse task of Logic2text. 
Each $d$ can have multiple  associated $\langle l, t \rangle$ pairs and each pair has a pre-defined logic type $c \in \mathcal{C}$ indicating its logical operation, where $\mathcal{C}$ = \{count, comparative, superlative, unique, ordinal, aggregation, majority\}. These logic types have different preferences on the patterns of logical forms (LFs) and texts. As shown in Figure \ref{fig:example}, \textit{count}-type LFs tend to contain \textit{eq} and \textit{count} functions; \textit{superlative}-type texts usually have superlative words like \textit{most, least, highest}, etc. For both tasks, we have the same supervision data $\mathcal{S}=\{(c_i,d_i,l_i,t_i)\}_{i=1}^k$, where $k$ is the number of instances.

\section{Proposed Approach}
Our framework is composed of three main stages. (1) Topic-conditioned data augmentation for augmenting logical forms and texts. (2) Round-trip data weighting for weighting the augmented data via round-trip reconstruction scores. (3) Joint training of Logic2text and LG, to utilize the augmented data to jointly train these models.
Figure \ref{fig:framework} depicts an overview of the framework.

\subsection{Base Model}
\label{sec:base}
\begin{figure}[t]
    \centering
    \includegraphics[width=0.9\linewidth]{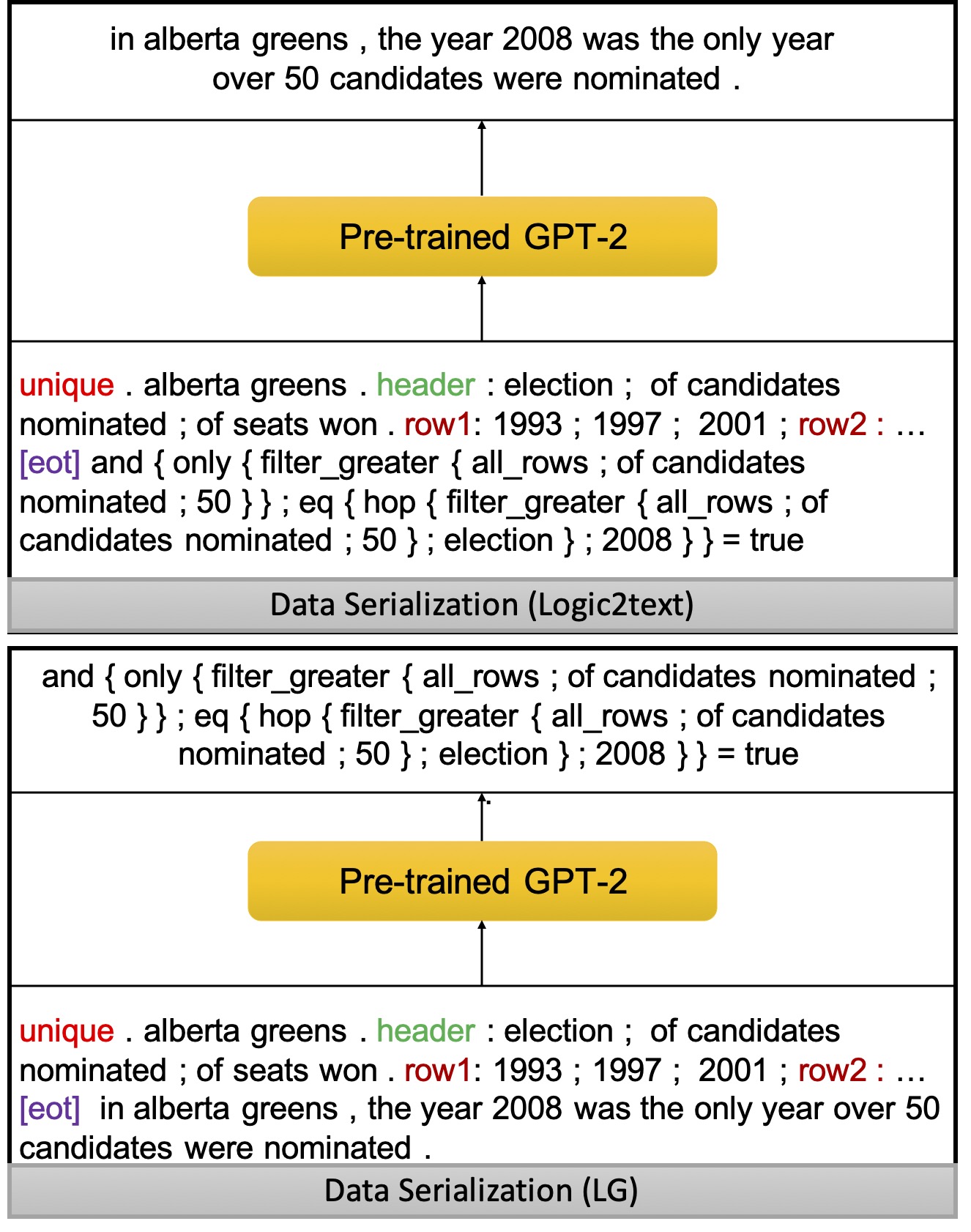}
    \caption{Base model for Logic2text and LG. The input of Logic2text is the concatenation of the logic type, table caption, table column headers, table content and linearized logical form. The output is the textual description. For LG, the positions of the logical form and text are switched.}
    \label{fig:base}
\end{figure}

Following \citet{chen2020logic2text}, we use a pre-trained GPT-2 model as the supervised base model. We employ the same data serialization as in \cite{chen2020logic2text} to represent the tables and logical forms as text sequences, as shown in Figure \ref{fig:base}.
Thus, each instance has a serialized table $d$ and logical form $l$. We additionally consider the logic type $c$ as an explicit prior knowledge to prompt the generation. Therefore, the objective of a Logic2text model is to generate a sequence $t^*=\{u_1, u_2, \dots, u_N\}$:
\begin{equation}
   t^* = \mathop{\arg\max}\prod_{i=1}^N P(u_i | u_{<i}, [c; d; l]; \theta),
    \label{eq:l2t-sup}
\end{equation}
where $[;]$ denotes concatenation of multiple sequences.
Similarly, for the LG task, the goal is to generate a logical form string $l^*=\{v_1, v_2, \dots, v_M\}$:

\begin{equation}
  l^* = \arg\max\prod_{j=1}^M P(v_j | v_{<j}, [c; d; t]; \phi).
    \label{eq:LG-sup}
\end{equation}

\subsection{Topic-Conditioned Data Augmentation}

Researchers have used pre-trained language models such as GPT-2 to augment text samples for text classification~\cite{papanikolaou2020dare,kumar2020data} and generation tasks~\cite{chang2021neural} by fine-tuning it on in-domain texts. The prior knowledge integrated in such pre-trained models leads to a considerable quality of the generated text. Different from previous work~\cite{chang2021jointly} that bootstraps new text instances from the original texts, we seek to generate logical forms and texts from tables.

To this end, we construct two \textit{conditional table-to-logic} ($P_{d2l}(l|d, c)$) and \textit{table-to-text} ($P_{d2t}(t|d, c)$) models, which require the generation of a logical form $l$ or text $t$ directly from a table $d$ following a certain logic type $c$. A logic type serves as a topic to control the generation pattern.
The objectives for these tasks are,
\begin{align}
    \mathcal{L}_{d2l} = \mathbb{E}_{(c, d, l) \sim \mathcal{S}}[-\log P_{d2l}(l| d, c)], \nonumber \\
    \mathcal{L}_{d2t} = \mathbb{E}_{(c, d, t) \sim \mathcal{S}}[-\log P_{d2t}(t| d, c)].
    \label{eq:sup}
\end{align}

The GPT-2 models fine-tuned on these tasks can be utilized for data augmentation through inference. Specifically, we perform inference with the trained table-to-logic and table-to-text models to generate extra LFs and texts based on a table $d_u$, where $d_u$ can be tables in the original training data or from other resources. We assign each of the seven logic types to $d_u$ to generate outputs with diverse logical patterns. As shown in Figure \ref{fig:framework}, the augmented data can be consistent with the assigned logic type.

A generated text or LF is filtered out if: (1) its length exceeds 200 tokens in Byte-Pair Encoding (BPE)~\cite{sennrich2016neural}; (2) it is identical to an existing instance in the training data.
In this work, we only consider the seven pre-defined logic types as topics. However, our method can be easily adapted to other domains and topics if it is provided with topic-dependent supervision data. For instance, we can use more fine-grained logic patterns like specific logical functions and words.

\subsection{Round-trip Data Weighting}
\label{sec:weight}
The augmented logical forms and texts can be noisy because they are generated from imperfect neural models. 
Therefore, we propose a round-trip data weighting (RT-Weight) strategy to assign per-example weights to the augmented data, which can balance their effects during unsupervised training. Particularly, we use a \textbf{round-trip BERTScore} metric. In round-trip translation, we first translate a logical form $l$ into a text using a Logic2text model and then back-translates it to a reconstructed $\tilde{l}$ with an LG model. We then compute the similarity between the $l$ and $\tilde{l}$ with BERTScore~\cite{bert-score} as the weight for this instance. The same method also applies to unlabeled text instances. Finally, we obtain a weight vector $W_l=[w_l^{(1)}, w_l^{(2)}, \dots, w_l^{(k_l)}]$ for $\mathcal{U}_l$ and $W_t=[w_t^{(1)}, w_t^{(2)},\dots, w_t^{(k_t)}]$ for $\mathcal{U}_t$, where $k_l$ and $k_t$ are the sizes of $\mathcal{U}_l$ and $\mathcal{U}_t$, respectively.

Our method is inspired by a line of works~\cite{imankulova2017improving,khatri2020filtering} on unsupervised machine translation, which use the round-trip BLEU~\cite{papineni2002bleu} score to filter out low-quality pseudo instances. However, a BLEU score relies on the overlapping n-grams between texts, not considering semantic-level similarity. Instead, we use the BERTScore, a popular evaluation metric for text generation, which leverages the pre-trained contextual embeddings from BERT~\cite{devlin2018bert} and compute cosine similarity between words in candidate and reference sentences. We adopt the F1-measure of BERTScore in practice. 
\subsection{Joint training of Logic2text and LG}
After data augmentation, we obtain $\mathcal{U}_l = \{(c_u, d_u, l_u)\}$ and $\mathcal{U}_t = \{(c_u, d_u, t_u)\}$, where $l_u$ and $t_u$ are unpaired logical forms and text descriptions, $d_u$ is a context table, and $c_u$ indicates the assigned logic type. They are not directly usable as supervised data for Logic2text and LG because $l_u$ and $t_u$ are generated independently and unaligned. Therefore, we aim to leverage these data to train models on both tasks.

Let $P_{\theta}(t|l,d)$ denote a Logic2text (L2T) model and $P_{\phi}(l|t,d)$ an LG model. Both models are first pre-trained on supervised data $\mathcal{S}$ for several epochs. We then employ unsupervised training as follows.

\paragraph{Back-translation}  Back-translation (BT) is commonly used in machine translation (MT)~\cite{edunov2018understanding} for augmenting pseudo-parallel data. Its core idea is to
translate a \textbf{target-side} monolingual sentence $y$ into a pseudo source language sentence $\hat{x}$, with a target-to-source translation model $M_{yx}$, which forms a pseudo parallel sentence $(\hat{x}, y)$ that can be used to train the source-to-target model $M_{xy}$. In our case, $P_{\theta}(t|l,d)$ and $P_{\phi}(l|t,d)$ are conditionally inverse to each other, because the table $d$ acts as a condition of the conversion between the logical form $l$ and text $t$. Therefore, we optimize the following back-translation objectives,
\begin{align}
 \mathcal{L}_{BT}(\theta) &= \mathbb{E}_{(d,t)\sim\mathcal{U}_t,w_t\sim W_t}[- w_t \log P_{\theta}(t|\hat{l},d)], \nonumber\\
    \mathcal{L}_{BT}(\phi) &=\mathbb{E}_{(d,l)\sim\mathcal{U}_l,w_l\sim W_l}[- w_l \log P_{\phi}(l|\hat{t},d)], \nonumber \\ 
    \hat{t}&\sim \arg\max_t P_{\theta}(t|l,d), \nonumber\\
    \hat{l}&\sim \arg\max_l  P_{\phi}(l|t,d),
    \label{eq:bt}
\end{align}
where $w_l$ and $w_t$ are the corresponding weights for the augmented data. 

\paragraph{Self-training} A drawback of back-translation is that it can only improve a model with its target-side unpaired data. In our case, the L2T model is trained only on the pseudo instances constructed from $U_t$ and the LG model only leverages pseudo data from $U_l$. To fully utilize the augmented data, we incorporate a self-training scheme, where each model predicts pseudo targets for its source-side unpaired data, constructing pseudo-parallel instances to re-train itself. These equations present self-training objectives.
\begin{align}
    \mathcal{L}_{ST}(\theta) &= \mathbb{E}_{(d,l)\sim\mathcal{U}_l,w_l\sim W_l}[- w_l \log P_{\theta}(\hat{t}|l,d)] \nonumber \\ 
    \mathcal{L}_{ST}(\phi)& = \mathbb{E}_{(d,t)\sim\mathcal{U}_t,w_t\sim W_t}[-  w_t \log P_{\phi}(\hat{l}|t,d)]  \nonumber \\
    \hat{t}&\sim \arg\max_t P_{\theta}(t|l,d) \nonumber\\
    \hat{l}&\sim \arg\max_l  P_{\phi}(l|t,d)
    \label{eq:st}
\end{align}

\paragraph{Curriculum learning} Furthermore, we adopt the curriculum learning~\cite{bengio2009curriculum} strategy by sorting the augmented data in descending order of their corresponding weights. The motivation is that the models can learn from easy augmented data (in which the generated pseudo examples are cleaner) to harder ones. This setting encourages the models to learn from pseudo-parallel data of higher quality in the beginning, and gradually transit to more error-prone ones that have lower weights.
We expect that this strategy can better balance the effects of the noisy augmented data.

\paragraph{Training scheme}
 We adopt a teacher-student training scheme to optimize the two models. At each epoch, we have both a teacher copy and a student copy of each model. During the unsupervised training, the teacher models are frozen to generate pseudo-parallel data for the student models. At the end of each epoch, the teacher models are updated from the corresponding student models. This is similar to the epoch-level \textit{Iterative Back-Translation} (IBT) scheme presented in \cite{xu2020dual,zhang2018joint}. We also perform teacher forcing, to fine-tune the models on the clean supervised data $\mathcal{S}$ at the end of each epoch. A formal description of the entire framework is illustrated in Appendix.

\section{Experiments}

\subsection{Datasets}
We conduct the main experiments on the Logic2text (L2T) dataset~\cite{chen2020logic2text}, a crowd-sourced logical-level NLG dataset containing 10.8k instances split into 8566/1095/1092 for train/val/test. The input of each instance is a database-like table and a logical form that describes a logical-level fact in the table, and the output is a text description. Each table is associated with 1 to 3 instances with different logic types. We reuse the L2T dataset to construct a logical form generation (LG) semantic parsing dataset, in which a logical form is generated from a table and a text description. LG follows the same train/val/test split as L2T. We experiment on (1) the full data setting, where all the supervised data are used for data augmentation and joint training, and (2) the few-shot setting, in which we randomly choose 1k training samples (according to the ratio of logic types in the original dataset) from the original supervision data for training DA models and joint training. However, we use all the tables in the original supervision data for data augmentation, which simulates the scenario where additional tables are incorporated. Table \ref{tab:aug} lists the statistics of the augmented data.

\subsection{Evaluation Metrics}
\label{sec:metric}
We use BLEU-4\footnote{multi-bleu.pl}, ROUGE-1,2,4,L\footnote{rouge-1.5.5} to evaluate the models on the Logic2text task, following ~\cite{chen2020logic2text}. For the LG task, we adopt the Logical Form Accuracy (LF Acc.) and Execution Accuracy (Exec. Acc.) metrics in a similar setting to a semantic parsing dataset WikiSQL~\cite{wikisql}. LF Acc. is the accuracy of the generated logical forms that have the exact string match with gold references. Exec. Acc. relaxes the criterion of LF Acc.: if the generated logical form can be successfully executed on the table, it counts for a correct prediction. Exec. Acc. has the downside that the generated logical form may not be consistent with the input text, but happens to be supported by the table.

\subsection{Model Configuration}
We implement our model on the Huggingface Transformers library~\cite{wolf-etal-2020-transformers} and PyTorch~\cite{paszke2019pytorch}. All experiments are conducted on NVIDIA 2080ti GPUs. We use GPT-2-small as our base model and Adam~\cite{kingma2014adam} optimizer. For base models without semi-supervised learning, we set the batch size to 4 and the learning rate to $3\times 10^{-5}$. For semi-supervised experiments, we set the batch size to $2$ and the learning rate to $2\times 10^{-5}$ because of the limit of the GPU memory. We employ beam search with the size of 3 for the decoding in both data augmentation and semi-supervised learning. The hyperparameters and best checkpoints of the Logic2text models are chosen based on the BLEU score on the validation set, and for LG, they are chosen based on Logical Form Accuracy on the validation set. Refer to Appendix \ref{sec:appendix} for more detail.

\begin{table}[t]
    \centering
    \resizebox{\columnwidth}{!}{
    \begin{tabular}{lccc}
    \toprule
      &  full data & few-shot data\\
      \midrule
     \# of instances for training DA models &  8566  &  1000  \\
     \# of tables for training DA models & 4554 & 929 \\
     \# of tables for DA inference & 4554  & 4554 \\
     \# of augmented logical forms & 29232 & 29253 \\
     \# of augmented texts  & 31548  & 31589  \\

    \bottomrule
    \end{tabular}}
    \caption{Statistics of data augmentation (DA). Note that we used all 4554 tables in the original training set for DA inference even for the few-shot setting.}
    \label{tab:aug}
\end{table}

\begin{table*}[t]
    \centering
    \small
    \begin{tabular}{lccccccc}
        \toprule
        & \multicolumn{5}{c}{\textbf{Logic2text}} & \multicolumn{2}{c}{\textbf{LG}} \\
        \quad {Models} & BLEU-4 & ROUGE-1 & ROUGE-2 & ROUGE-4 &ROUGE-L & LF Acc. & Exec. Acc. \\
        \midrule
        
        Template & 17.57 & 50.56 &24.20& 6.61 &37.81 & -- & --\\
        Seq2seq+att & 12.46 &36.22 &15.91 &4.49 &31.03 & -- &-- \\
        Pointer generator & 24.03 &56.23 &30.51& 10.78& 46.85 & -- & -- \\
        Graph2seq+copy & 25.38 & 58.15& 32.79& 12.25& 49.47& -- &--\\
        Transformer+copy &26.42& 58.77& 33.05& 12.83& 49.01&--&-- \\
        GPT-2~\cite{chen2020logic2text} &31.44 &64.16 &39.48 &17.46& 53.99& -- & -- \\
        \midrule
        \textit{Our Implementations }\\

        GPT-2 (full)& 31.69&	64.88&	40.77	&18.29	&54.89& 67.03 & 88.74 \\
     
         Ours (full) & \textbf{32.68}	& \textbf{65.74} &	\textbf{41.54}	&\textbf{19.16} &\textbf{	55.50}	&\textbf{67.95 }& \textbf{89.01} \\
        
           GPT-2 (1k) & 23.38 & 56.65 & 31.19  &11.03 & 46.39 & 41.30 & 69.60 \\
          Ours (1k) & 26.34	& 59.79 &34.86	&14.08 &	49.18	& 45.51 & 71.98\\
  
        \bottomrule
    \end{tabular}
    \caption{Main results on the test splits of Logic2text and LG.}
    \label{tab:res}
\end{table*}

\subsection{Models for Comparison} 
We compare the proposed method with the previous supervised baselines~\cite{chen2020logic2text} on Logic2text. \textbf{Template}: Manually crafted generation templates for different logic types based on logical forms. \textbf{Seq2seq+att}: This is an adapted seq2seq model with an attention mechanism. \textbf{Pointer generator}: A copy mechanism is added to Seq2seq+att to allow the model to copy from the inputs, which is crucial for the fidelity-preserving NLG tasks with abundant entity names and values. \textbf{Graph2seq+copy}: Graph2seq~\cite{xu2018sql} builds a graph-based encoder to encode the logical forms with the copy mechanism. \textbf{Transformer+copy}: This approach is based on the vanilla Transformer~\cite{vaswani2017attention} network with extra copy mechanism. \textbf{GPT-2}: A pre-trained GPT-2 model is fine-tuned on the Logic2text, where the input tables and logical forms are represented as text sequences. This is the state-of-the-art of Logic2text. Moreover, we re-implement the GPT-2 model with the Huggingface Transformers library, showing higher training speed than theirs. We adapt the GPT-2 model to both Logic2text and LG. As described in Section \ref{sec:base}, we add logic types as additional inputs in our GPT-2 implementation.

\subsection{Main Results}
\label{sec:main}
Table \ref{tab:res} summarizes the main results on the test splits of Logic2text and LG. We can observe that the models based on pre-trained GPT-2 completely outperform previous neural models without pre-training. The GPT-2 model we implemented obtains generally better results than the GPT-2 model implemented by \citet{chen2020logic2text} on all the metrics of Logic2text.
This demonstrates the additional merit of using logic type information. Our full model with augmented data outperforms the base model by almost 1 BLEU score and more than 0.5 points on all ROUGE scores. 

However, our full model has relatively trivial improvements over the base model on Logical Form Accuracy (LF. Acc.) and Execution Accuracy (Exer. Acc.). This is probably because LG is not as difficult as Logic2text and the full supervised data are already enough for training a powerful LG model. Moreover, we evaluated the models of GPT-2 and Ours on the few-shot subset of 1k training data. The results of Ours (1k) are remarkably better than GPT-2 (1k) on all metrics of Logic2text and LG, indicating the particular effects of our method in low-resource scenarios. It is worth noting that GPT-2 is already a powerful model pre-trained on large-scale corpora and that our approach can offer additional improvements with augmented in-domain data. 
\begin{table*}[t]
    \centering
    \small
    \begin{tabular}{lccccccc}

        \toprule
        & \multicolumn{5}{c}{\textbf{Logic2text}} & \multicolumn{2}{c}{\textbf{LG}} \\
        \quad {Models} & BLEU-4 & ROUGE-1 & ROUGE-2 & ROUGE-4 &ROUGE-L & LF Acc. & Exec. Acc. \\
        \midrule
        Full model & \textbf{32.68}	& \textbf{65.74 }&	\textbf{41.54}	&\textbf{19.16 }&\textbf{	55.50}	&\textbf{67.95} & 89.01 \\
        \midrule
        -- ST & 32.53 &	65.35 &	41.23 &	18.82 &	55.16&	67.40 & \textbf{89.38} \\
        -- BT & 31.74. & 64.67 & 40.60 & 18.45 & 54.78 & 64.56 &  87.09\\
        -- order & 32.48 & 65.21 & 41.45 & 19.11 & 55.14 & 66.67 &87.73 \\
        -- weight & 31.93 &	65.02 &	40.60 &	18.49 &	54.45 & 	66.85 &		89.19 \\
        
        \bottomrule
    \end{tabular}
    \caption{Ablation results on the test sets of Logic2text and LG.}
    \label{tab:abl}
\end{table*}
\subsection{Ablation study}
To validate the effectiveness of our joint training method, we conduct ablation studies on four ablated variants of the full model with full supervised data: (i) --ST: remove the self-training part in Equation \eqref{eq:st}; (ii) --BT: remove the back-translation part in Equation \eqref{eq:bt}; (iii) --order: remove the curriculum learning setting; (iv) --weight: remove data weighting and curriculum learning, which means treating all augmented data equally.

\paragraph{Logic2text results} As shown in Table \ref{tab:abl}, removing any of the components causes a drop of performance on Logic2text. Particularly, removing BT drastically hurts the performance, making it unable to compete the base model. It is reasonable because removing BT loses interactions between the Logic2text and LG models and they are only trained separately via self-training. This observation implies the importance of the joint training of Logic2text and LG. We also observe that self-training does not contribute to the final result as much as back-translation does. This is reasonable as self-training is based on the prediction of each model itself, making the model apt to learn its own mistakes, while BT enables each model to learn from supervision signals produced by an opposite model. Moreover, the curriculum learning setting also consistently contributes to the performance. The data weighting is also essential for achieving the best performance.

\paragraph{LG results} All the components contribute to the LF Acc., whereas BT is the most important component. However, removing ST or weighting enhances Exec. Acc., possibly because Exec. Acc. is an approximated metric as defined in Section \ref{sec:metric}. The observations suggest the need for better evaluation metrics for LG.

\subsection{Human Evaluation}
\label{sec:humaneval}
 We conduct human evaluation to further test the quality of Logic2text generation. We randomly sampled 100 instances from the generations of four models as described in Section \ref{sec:main}: (1) GPT-2 (full), (2) Ours (full), (3) GPT-2 (1k) and (4) Ours (1k), along with (5) the gold references. We follow \citet{chen2020logic2text} to evaluate on two metrics: (1) \textit{factual correctness}, i.e., whether the generated description is factually supported by the table, which is also referred to as logical fidelity; (2) \textit{semantic correctness}, i.e., whether the generated description is consistent with the meaning of the logical form. We ask 3 human experts (computer science graduate students who are familiar with the logical form schema) to evaluate each example and take votes of the results as the final decisions, i.e., a sample is judged as correct only if at least two people agree with it.
 
 We present the accuracies of both metrics in Table \ref{tab:humaneval}. As can be observed, our methods outperform the base model GPT-2 under both full-data and few-shot settings, which is generally consistent with the Automatic Evaluation results. In particular, Ours (1k) outperforms GPT-2 (1k) on both metrics by over 10\%.
 
\begin{table}[t]
    \centering
    \begin{tabular}{lcc}
    
    \toprule
      Models   & Factual Acc. & Semantical Acc. \\
      \midrule

     Gold    &  0.99 & 0.95 \\
    \midrule
    GPT-2 (full) & 0.84 & 0.74 \\
    Ours (full) & 0.90 & 0.77 \\
    \midrule
    GPT-2 (1k) & 0.51 & 0.39 \\
    Ours (1k) & 0.67 & 0.53 \\
    
    \bottomrule
    \end{tabular}
    \caption{Human evaluation on 100 sampled instances from the test set of Logic2text. }
    \label{tab:humaneval}
\end{table}

\subsection{Quality of Data Augmentation}
\label{sec:quality}
Here, we analyze whether TopicDA can generate high-quality data, based on two metrics evaluated on the augmented data in the full-data setting. 

\paragraph{Topic Consistency} The main motivation of our TopicDA method is to use logic types as topics to encourage the DA models to generate topic-diversified data. Therefore, we analyze whether the augmented data are consistent with the pre-assigned topic. To realize this evaluation, we train two auxiliary topic classifiers LF-CLR and Text-CLR for classifying the logic type of a logical form and a textual description, respectively. The two classifiers are pre-trained BERT-base~\cite{devlin2018bert} models fine-tuned on the training set of Logic2text and then validated on the test set. The validation accuracy of LF-CLR reaches 100\% and the Text-CLR achieves 97.7\%. These classifier models are then used to evaluate whether the augmented data are consistent with their assigned logic types during augmentation. The evaluation results listed in Table \ref{tab:quality} suggest that the augmented data are generally topically-consistent with accuracies of 98.52\% and 94.30\% for LFs and texts, respectively. This demonstrates the effectiveness of our TopicDA model.

\paragraph{Factual Correctness} The augmented logical forms and texts can be noisy as we described in Section \ref{sec:weight}. Similarly to the factual correctness accuracy defined in Section \ref{sec:humaneval}, we validate whether the augmented data can be exactly supported by their assigned tables. For logical forms, we directly execute them on the table and compute the execution accuracy. For texts, we back-translate them to logical forms with an LG model and then execute the logical forms on the tables. The accuracy of factual correctness is 11.75\% and 21.03\% for logical forms and texts, respectively. This result is reasonable because the factual correctness is evaluated with the exact match of the logical form execution, which allows no ambiguity in the natural language arguments.  
\begin{table}[t]
    \centering
    
    \begin{tabular}{lcc}
    \toprule
         & Topic Acc.  & Factual Acc. \\
         \midrule
     Logical forms    & 98.52\% &  11.75\% \\
     Texts & 94.30\% & 21.03\% \\
     \bottomrule
    \end{tabular}
    \caption{Evaluation on the quality of the augmented data.}
    \label{tab:quality}
\end{table}

\paragraph{Validity of Logical Forms} We also test the validity of the generated logical forms. Out of the 29232 augmented logical forms (string-type) in the full-data setting, 27,648 (94.58\%) can be parsed into valid logical forms without explicit structural errors (e.g., incorrect functions, misplaced punctuations, and mismatch between the number of arguments and functions); 15,951 (54.57\%) can be successfully executed on the corresponding tables regardless of the correctness of the outputs, i.e., we can obtain a Boolean result for a logical form, but the result may be correct or wrong. This demonstrates the effectiveness of our seq2seq model for logical form generation. Although the logical forms may not be factually supported by the context tables, we still find them beneficial to the logic-text conversion because of their general validity. Some qualitative examples of TopicDA are provided in Appendix \ref{sec:aug-eg}.

\subsection{Effects of LG for Annotation}
In this experiment, we analyze whether the LG task is beneficial to the annotation work of new logical-level NLG datasets. We choose the LogicNLG ~\cite{chen2020logicalNLG} dataset, which aims to generate logical descriptions from tables without any intermediate logical forms. Neural models perform poorly on this dataset owing to the low fidelity and uncontrollable content selection, as described in Section \ref{sec:intro}. We use the LG model that we built to generate logical forms (LFs) based on the tables and textual statements in LogicNLG.\footnote{ Note that although Logic2text and LogicNLG have many common tables, the texts in Logic2text are separately annotated and do not overlap with those of LogicNLG.} We adopt the same GPT-TabGen (sm) model in \cite{chen2020logicalNLG}, with the generated LFs as additional inputs similarly to the case of the Logic2text.
The model with additional LFs shows substantial improvements on LogicNLG, as shown in Table \ref{tab:logicnlg}. The results are not exactly comparable, because we introduce extra LFs for both training and test sets. However, they still reveal that such intermediate logical forms play an important role for a better logical NLG system. We expect future work on utilizing LG for annotating benchmark data.

\begin{table}[t]
    \centering
    \resizebox{\columnwidth}{!}{
    \begin{tabular}{lccc}
    \toprule
         & BLEU-1 & BLEU-2 &BLEU-3  \\
         \midrule
    GPT-TabGen (sm)     &  48.8 & 27.1 &12.6 \\
    GPT-TabGen (sm) + LFs & \textbf{55.8} & \textbf{35.6} & \textbf{20.5} \\ 
    \bottomrule
    \end{tabular}}
    \caption{Results on the test split of LogicNLG.}
    \label{tab:logicnlg}
\end{table}

\section{Related Work}

 Back-translation is a popular method for semi-supervised natural language generation, which has been proven effective in machine translation~\cite{edunov2018understanding}. Iterative back-translation (IBT)~\cite{hoang2018ibt, guo2020revisiting} is an extension of back-translation, in which forward and backward models are trained together to generate pseudo parallel instances with each other. A similar line of studies~\cite{su2020towards,tang2017question} adopts dual learning, which incorporates the inference process of both models into training via reinforcement learning. IBT and dual learning are based on the same idea, to jointly solve tasks with duality used in machine translation. This idea is currently popular in data-to-text generation~\cite{chang2021neural, chang2021jointly, qader2019semi, guo2020cyclegt, schmitt2020unsupervised}. Our work adopts a similar idea to model the conditional duality between Logic2text~\cite{chen2020logic2text} and Logical Form Generation. We adopt an IBT-style joint training scheme without the back propagation through inference models. We also incorporate a self-training strategy to combine the advantage of both sides of unpaired data.
Different from previous works~\cite{qader2019semi, su2020towards, guo2020cyclegt} that directly exploited off-the-shelf unpaired data, \citet{chang2021neural} posed the problem that it is unrealistic to have so much unpaired texts for data-to-text tasks, and adopt language models (LM), i.e., GPT-2~\cite{radford2019gpt2} to augment additional texts. The Logic2text task further toughens the problem because we must augment logical forms and textual statements supported by the tables, suggesting the difficulty to apply dual learning. In this work, we realize the data augmentation from existing tables without the availability of additional resources.
Our work is also related to \cite{dou2020dynamic} that provides useful insights for data selection in IBT. Our work adopts a round-trip BERTScore metric that can measure the quality of both texts and logical forms.

\section{Conclusion}
We studied the logical-level NLG task with the limited supervision data. We herein proposed a topic-conditioned data augmentation method to generate logical forms and textual descriptions with GPT-2. We also introduced logical form generation as a dual task of logical-level NLG, and propose a joint semi-supervised learning approach to improve these two tasks with augmented data. The experimental results show the effectiveness of the proposed method, especially in low-resource settings. For future work, we seek to apply our method to annotate new logical-level NLG benchmarks.

\section*{Acknowledgements}
 This paper is based on results obtained from a project, JPNP18002, commissioned by the New Energy and Industrial Technology Development Organization (NEDO).

\bibliography{custom}

\begin{thebibliography}{34}
\providecommand{\natexlab}[1]{#1}

\bibitem[{Bengio et~al.(2009)Bengio, Louradour, Collobert, and
  Weston}]{bengio2009curriculum}
Bengio, Y.; Louradour, J.; Collobert, R.; and Weston, J. 2009.
\newblock Curriculum learning.
\newblock In \emph{Proceedings of the 26th annual international conference on
  machine learning}, 41--48.

\bibitem[{Chang, Demberg, and Marin(2021)}]{chang2021jointly}
Chang, E.; Demberg, V.; and Marin, A. 2021.
\newblock Jointly Improving Language Understanding and Generation with
  Quality-Weighted Weak Supervision of Automatic Labeling.
\newblock In \emph{Proceedings of the 16th Conference of the European Chapter
  of the Association for Computational Linguistics: Main Volume}, 818--829.

\bibitem[{Chang et~al.(2021)Chang, Shen, Zhu, Demberg, and
  Su}]{chang2021neural}
Chang, E.; Shen, X.; Zhu, D.; Demberg, V.; and Su, H. 2021.
\newblock Neural Data-to-Text Generation with LM-based Text Augmentation.
\newblock In \emph{Proceedings of the 16th Conference of the European Chapter
  of the Association for Computational Linguistics: Main Volume}, 758--768.

\bibitem[{Chen et~al.(2020{\natexlab{a}})Chen, Chen, Su, Chen, and
  Wang}]{chen2020logicalNLG}
Chen, W.; Chen, J.; Su, Y.; Chen, Z.; and Wang, W.~Y. 2020{\natexlab{a}}.
\newblock Logical Natural Language Generation from Open-Domain Tables.
\newblock In \emph{Proceedings of the 58th Annual Meeting of the Association
  for Computational Linguistics}, 7929--7942.

\bibitem[{Chen et~al.(2020{\natexlab{b}})Chen, Chen, Zha, Zhou, Zhang,
  Sundaresan, and Wang}]{chen2020logic2text}
Chen, Z.; Chen, W.; Zha, H.; Zhou, X.; Zhang, Y.; Sundaresan, S.; and Wang,
  W.~Y. 2020{\natexlab{b}}.
\newblock Logic2Text: High-Fidelity Natural Language Generation from Logical
  Forms.
\newblock In \emph{Proceedings of the 2020 Conference on Empirical Methods in
  Natural Language Processing: Findings}, 2096--2111.

\bibitem[{Devlin et~al.(2018)Devlin, Chang, Lee, and
  Toutanova}]{devlin2018bert}
Devlin, J.; Chang, M.-W.; Lee, K.; and Toutanova, K. 2018.
\newblock Bert: Pre-training of deep bidirectional transformers for language
  understanding.
\newblock \emph{arXiv preprint arXiv:1810.04805}.

\bibitem[{Dou, Anastasopoulos, and Neubig(2020)}]{dou2020dynamic}
Dou, Z.-Y.; Anastasopoulos, A.; and Neubig, G. 2020.
\newblock Dynamic Data Selection and Weighting for Iterative Back-Translation.
\newblock In \emph{Proceedings of the 2020 Conference on Empirical Methods in
  Natural Language Processing (EMNLP)}, 5894--5904.

\bibitem[{Edunov et~al.(2018)Edunov, Ott, Auli, and
  Grangier}]{edunov2018understanding}
Edunov, S.; Ott, M.; Auli, M.; and Grangier, D. 2018.
\newblock Understanding Back-Translation at Scale.
\newblock In \emph{Proceedings of the 2018 Conference on Empirical Methods in
  Natural Language Processing}, 489--500.

\bibitem[{Guo et~al.(2020)Guo, Jin, Qiu, Zhang, Wipf, and
  Zhang}]{guo2020cyclegt}
Guo, Q.; Jin, Z.; Qiu, X.; Zhang, W.; Wipf, D.; and Zhang, Z. 2020.
\newblock CycleGT: Unsupervised Graph-to-Text and Text-to-Graph Generation via
  Cycle Training.
\newblock \emph{CoRR}, abs/2006.04702.

\bibitem[{Guo et~al.(2021)Guo, Zhu, Lin, Chen, Lou, and
  Zhang}]{guo2020revisiting}
Guo, Y.; Zhu, H.; Lin, Z.; Chen, B.; Lou, J.-G.; and Zhang, D. 2021.
\newblock Revisiting Iterative Back-Translation from the Perspective of
  Compositional Generalization.

\bibitem[{Hoang et~al.(2018)Hoang, Koehn, Haffari, and Cohn}]{hoang2018ibt}
Hoang, V. C.~D.; Koehn, P.; Haffari, G.; and Cohn, T. 2018.
\newblock Iterative back-translation for neural machine translation.
\newblock In \emph{Proceedings of the 2nd Workshop on Neural Machine
  Translation and Generation}, 18--24.

\bibitem[{Imankulova, Sato, and Komachi(2017)}]{imankulova2017improving}
Imankulova, A.; Sato, T.; and Komachi, M. 2017.
\newblock Improving low-resource neural machine translation with filtered
  pseudo-parallel corpus.
\newblock In \emph{Proceedings of the 4th Workshop on Asian Translation
  (WAT2017)}, 70--78.

\bibitem[{Khatri and Bhattacharyya(2020)}]{khatri2020filtering}
Khatri, J.; and Bhattacharyya, P. 2020.
\newblock Filtering Back-Translated Data in Unsupervised Neural Machine
  Translation.
\newblock In \emph{Proceedings of the 28th International Conference on
  Computational Linguistics}, 4334--4339.

\bibitem[{Kingma and Ba(2014)}]{kingma2014adam}
Kingma, D.~P.; and Ba, J. 2014.
\newblock Adam: A method for stochastic optimization.
\newblock \emph{arXiv preprint arXiv:1412.6980}.

\bibitem[{Kumar, Choudhary, and Cho(2020)}]{kumar2020data}
Kumar, V.; Choudhary, A.; and Cho, E. 2020.
\newblock Data Augmentation using Pre-trained Transformer Models.
\newblock In \emph{Proceedings of the 2nd Workshop on Life-long Learning for
  Spoken Language Systems}, 18--26.

\bibitem[{Lebret, Grangier, and Auli(2016)}]{lebret2016wikibio}
Lebret, R.; Grangier, D.; and Auli, M. 2016.
\newblock Neural Text Generation from Structured Data with Application to the
  Biography Domain.
\newblock In \emph{Proceedings of the 2016 Conference on Empirical Methods in
  Natural Language Processing}, 1203--1213.

\bibitem[{Novikova, Du{\v{s}}ek, and Rieser(2017)}]{novikova2017e2e}
Novikova, J.; Du{\v{s}}ek, O.; and Rieser, V. 2017.
\newblock The E2E Dataset: New Challenges For End-to-End Generation.
\newblock In \emph{Proceedings of the 18th Annual SIGdial Meeting on Discourse
  and Dialogue}, 201--206.

\bibitem[{Papanikolaou and Pierleoni(2020)}]{papanikolaou2020dare}
Papanikolaou, Y.; and Pierleoni, A. 2020.
\newblock Dare: Data augmented relation extraction with gpt-2.
\newblock \emph{arXiv preprint arXiv:2004.13845}.

\bibitem[{Papineni et~al.(2002)Papineni, Roukos, Ward, and
  Zhu}]{papineni2002bleu}
Papineni, K.; Roukos, S.; Ward, T.; and Zhu, W.-J. 2002.
\newblock Bleu: a method for automatic evaluation of machine translation.
\newblock In \emph{Proceedings of the 40th annual meeting of the Association
  for Computational Linguistics}, 311--318.

\bibitem[{Paszke et~al.(2019)Paszke, Gross, Massa, Lerer, Bradbury, Chanan,
  Killeen, Lin, Gimelshein, Antiga et~al.}]{paszke2019pytorch}
Paszke, A.; Gross, S.; Massa, F.; Lerer, A.; Bradbury, J.; Chanan, G.; Killeen,
  T.; Lin, Z.; Gimelshein, N.; Antiga, L.; et~al. 2019.
\newblock Pytorch: An imperative style, high-performance deep learning library.
\newblock \emph{arXiv preprint arXiv:1912.01703}.

\bibitem[{Qader, Portet, and Labb{\'e}(2019)}]{qader2019semi}
Qader, R.; Portet, F.; and Labb{\'e}, C. 2019.
\newblock Semi-Supervised Neural Text Generation by Joint Learning of Natural
  Language Generation and Natural Language Understanding Models.
\newblock In \emph{Proceedings of the 12th International Conference on Natural
  Language Generation}, 552--562.

\bibitem[{Radford et~al.(2019)Radford, Wu, Child, Luan, Amodei, and
  Sutskever}]{radford2019gpt2}
Radford, A.; Wu, J.; Child, R.; Luan, D.; Amodei, D.; and Sutskever, I. 2019.
\newblock Language Models are Unsupervised Multitask Learners.

\bibitem[{Schmitt et~al.(2020)Schmitt, Sharifzadeh, Tresp, and
  Sch{\"u}tze}]{schmitt2020unsupervised}
Schmitt, M.; Sharifzadeh, S.; Tresp, V.; and Sch{\"u}tze, H. 2020.
\newblock An unsupervised joint system for text generation from knowledge
  graphs and semantic parsing.
\newblock In \emph{Proceedings of the 2020 Conference on Empirical Methods in
  Natural Language Processing (EMNLP)}, 7117--7130.

\bibitem[{Sennrich, Haddow, and Birch(2016)}]{sennrich2016neural}
Sennrich, R.; Haddow, B.; and Birch, A. 2016.
\newblock Neural Machine Translation of Rare Words with Subword Units.
\newblock In \emph{Proceedings of the 54th Annual Meeting of the Association
  for Computational Linguistics (Volume 1: Long Papers)}, 1715--1725.

\bibitem[{Su, Huang, and Chen(2020)}]{su2020towards}
Su, S.-Y.; Huang, C.-W.; and Chen, Y.-N. 2020.
\newblock Towards Unsupervised Language Understanding and Generation by Joint
  Dual Learning.
\newblock In \emph{Proceedings of the 58th Annual Meeting of the Association
  for Computational Linguistics}, 671--680.

\bibitem[{Tang et~al.(2017)Tang, Duan, Qin, Yan, and Zhou}]{tang2017question}
Tang, D.; Duan, N.; Qin, T.; Yan, Z.; and Zhou, M. 2017.
\newblock Question answering and question generation as dual tasks.
\newblock \emph{arXiv preprint arXiv:1706.02027}.

\bibitem[{Vaswani et~al.(2017)Vaswani, Shazeer, Parmar, Uszkoreit, Jones,
  Gomez, Kaiser, and Polosukhin}]{vaswani2017attention}
Vaswani, A.; Shazeer, N.; Parmar, N.; Uszkoreit, J.; Jones, L.; Gomez, A.~N.;
  Kaiser, {\L}.; and Polosukhin, I. 2017.
\newblock Attention is all you need.
\newblock In \emph{Proceedings of the 31st International Conference on Neural
  Information Processing Systems}, 6000--6010.

\bibitem[{Wenqing et~al.(2021)Wenqing, Jidong, Yitian, Hao, and
  Yaohui}]{deconfounded}
Wenqing, C.; Jidong, T.; Yitian, L.; Hao, H.; and Yaohui, J. 2021.
\newblock De-Confounded Variational Encoder-Decoder for Logical Table-to-Text
  Generation.
\newblock In \emph{Proceedings of the 59th Annual Meeting of the Association
  for Computational Linguistics}, 5532--5542.

\bibitem[{Wolf et~al.(2020)Wolf, Debut, Sanh, Chaumond, Delangue, Moi, Cistac,
  Rault, Louf, Funtowicz, Davison, Shleifer, von Platen, Ma, Jernite, Plu, Xu,
  Scao, Gugger, Drame, Lhoest, and Rush}]{wolf-etal-2020-transformers}
Wolf, T.; Debut, L.; Sanh, V.; Chaumond, J.; Delangue, C.; Moi, A.; Cistac, P.;
  Rault, T.; Louf, R.; Funtowicz, M.; Davison, J.; Shleifer, S.; von Platen,
  P.; Ma, C.; Jernite, Y.; Plu, J.; Xu, C.; Scao, T.~L.; Gugger, S.; Drame, M.;
  Lhoest, Q.; and Rush, A.~M. 2020.
\newblock Transformers: State-of-the-Art Natural Language Processing.
\newblock In \emph{Proceedings of the 2020 Conference on Empirical Methods in
  Natural Language Processing: System Demonstrations}, 38--45. Online:
  Association for Computational Linguistics.

\bibitem[{Xu et~al.(2018)Xu, Wu, Wang, Feng, and Sheinin}]{xu2018sql}
Xu, K.; Wu, L.; Wang, Z.; Feng, Y.; and Sheinin, V. 2018.
\newblock SQL-to-Text Generation with Graph-to-Sequence Model.
\newblock In \emph{Proceedings of the 2018 Conference on Empirical Methods in
  Natural Language Processing}, 931--936.

\bibitem[{Xu, Niu, and Carpuat(2020)}]{xu2020dual}
Xu, W.; Niu, X.; and Carpuat, M. 2020.
\newblock Dual Reconstruction: a Unifying Objective for Semi-Supervised Neural
  Machine Translation.
\newblock In \emph{Proceedings of the 2020 Conference on Empirical Methods in
  Natural Language Processing: Findings}, 2006--2020.

\bibitem[{Zhang et~al.(2020)Zhang, Kishore, Wu, Weinberger, and
  Artzi}]{bert-score}
Zhang, T.; Kishore, V.; Wu, F.; Weinberger, K.~Q.; and Artzi, Y. 2020.
\newblock BERTScore: Evaluating Text Generation with BERT.
\newblock In \emph{International Conference on Learning Representations}.

\bibitem[{Zhang et~al.(2018)Zhang, Liu, Li, Zhou, and Chen}]{zhang2018joint}
Zhang, Z.; Liu, S.; Li, M.; Zhou, M.; and Chen, E. 2018.
\newblock Joint training for neural machine translation models with monolingual
  data.
\newblock In \emph{Proceedings of the AAAI Conference on Artificial
  Intelligence}, volume~32.

\bibitem[{Zhong, Xiong, and Socher(2017)}]{wikisql}
Zhong, V.; Xiong, C.; and Socher, R. 2017.
\newblock Seq2SQL: Generating Structured Queries from Natural Language using
  Reinforcement Learning.
\newblock \emph{CoRR}, abs/1709.00103.

\end{thebibliography}


\appendix
\section{Configuration Details of Models}
\label{sec:appendix}
Here we provide detailed configurations of our models:
our implementation is based on Huggingface Transformer v3.1.0's GPT-2-small model. The word embeddings and positional embeddings are frozen during training.
The input sequence, i.e., concatenation of a logic type, caption, headers, table content and logical form, has a maximum of 800 BPE tokens, in which the maximum length for table content is 400, and 200 for logical form and 50 for textual descriptions. If the maximum lengths of a batch of data is smaller than the pre-defined ones, they are set to the smaller ones. 

We experimented on a set of combinations of batch size (bs) and learning rate (lr) by manually tuning on the validation set results. With trial and error on bs=\{1,2,4,8,20\} and lr=[1e-5 :3e-4] with step size 1e-5. 
we found (bs=2, lr=2e-5), (bs=4,lr=3e-5), (bs=8,lr=5e-5) consistently good. We adopt (bs=2, lr=2e-5) for the full model due to the extra memory cost of pseudo data generation.
We use a gradient clipping with L2-norm and threshold 5.0. We fix the random seed randomly to 42 for all experiments. Because of the large experimental cost of semi-supervised learning, we did not perform multiple runs on all models. In Table \ref{tab:run}, we show the results based on 5 runs of random seed (42, 10, 11, 101, 111) to compare Ours(full) and Ours(1k).
\begin{table}[t]
    \centering
    \small
    \resizebox{\columnwidth}{!}{\begin{tabular}{lcccc}

        \toprule
        & \multicolumn{2}{c}{\textbf{Logic2text}} & \multicolumn{2}{c}{\textbf{LG}} \\
        \quad {Models} & BLEU-4 &ROUGE-L & LF Acc. & Exec. Acc. \\
        \midrule
        GPT-2 (full) & $31.60\pm 0.31$ & $54.45\pm 0.40$ & $66.70\pm 0.48$ & $88.43\pm 0.34$ \\
        Ours (full) &$32.70\pm 0.44$ & $55.08\pm 0.36$ & $67.27\pm 0.72$ & $89.34\pm 0.53$ \\
        \bottomrule
    \end{tabular}}
    \caption{Results on the test sets of Logic2text and LG, where mean and standard deviation are computed on 5 runs .}
    \label{tab:run}
\end{table}

For the full data setting, we pre-train our joint models on supervised data for 1 epoch and fine-tune for 1 epoch at the end of each unsupervised training epoch. For the few-shot setting, since supervised data are sparse, we pre-train for 5 epochs and fine-tune for 3 epochs. On an NVIDIA 2080ti GPU, it takes about 8 hours to finish one unsupervised epoch and around 30 min for a pre-training/fine-tuning epoch, including the time of validation.

\section{Effectiveness of Round-trip Data Weighting}
In Section \ref{sec:quality}, we measured the factual accuracy of the augmented data, which showed 11.75\% of the logical forms and 21.03\% of the texts are logically supported by the tables. However, we adopted a soft data weighting approach instead of hard data filtering based on factual correctness. This is because factual correctness is evaluated based on exact match based execution of the augmented logical forms and the back-translated logical forms from the augmented texts. Therefore, it may strictly filter out some logically correct examples. Additionally, even if the logical forms and texts are not supported by the tables, they may still help models learn the conversion between LFs and texts. In comparison, our round-trip data weighting method tend to assign lower scores to low-quality augmented data while still considering their potential effects. To analyse whether round-trip data weighting can measure the quality of augmented data, we split the augmented dataset ("All") into "Correct" and "Incorrect" based on their exact factual correctness. In Table \ref{tab:weight}, we show that "correct" data tend to have higher round-trip weights.
\begin{table}[t]
    \centering
    \begin{tabular}{c c cc}
    \toprule
     & All & Correct & Incorrect \\
     \midrule
       Texts  & 0.65±0.20 &	0.70±0.19 &	0.64±0.20 \\
       Logical forms  & 0.86±0.16 &	0.92±0.12 &	0.87±0.13 \\
       \bottomrule
    \end{tabular}
    \caption{Mean and standard deviations of the data weights for different subsets of the augmented data.}
    \label{tab:weight}
\end{table}

Empirically, we also find it beneficial to use more soft-filtered data instead of less hard-filtered data. We show some qualitative examples in Appendix \ref{sec:aug-eg} to present the potential effects of imperfect augmented data.

\section{A Formal Algorithm of Proposed Method}
\begin{algorithm}[ht]
  \caption{Training procedure}
  \label{alg}
\begin{algorithmic}[1]
\State{\bfseries Input:} Augmented logical form dataset $\mathcal{U}_l$; augmented text dataset $\mathcal{U}_t$; Supervised dataset $\mathcal{S}$; 
\State {\bfseries Output:} A Logic2text model $P_{\theta}$; an LG model $P_{\phi}$.

{\Comment{Pre-training}}
\State Initialize $P_{\theta}$ and $P_{\phi}$ with pre-trained GPT-2.
\State Train $P_{\theta}$ and $P_{\phi}$ on $\mathcal{S}$.

{\Comment{Data weighting}}
\State Compute data weights $W_l$ for $\mathcal{U}_l$ and $W_t$ for $\mathcal{U}_t$ via round-trip BERTScore. 
\State Sort $\mathcal{U}_l$ and $\mathcal{U}_t$ in the decreasing order of $W_l$ and $W_t$, respectively.

{\Comment{Joint training }}
\Repeat
\State Optimize $P_{\theta}$ and $P_{\phi}$ via back-translation on $\mathcal{U}_l$ and $\mathcal{U}_t$, according to Equation \eqref{eq:bt}.
\State Optimize $P_{\theta}$ and $P_{\phi}$ via self-training on $\mathcal{U}_l$ and $\mathcal{U}_t$, according to Equation \eqref{eq:st}.
\State Fine-tuning $P_{\theta}$ and $P_{\phi}$ on $\mathcal{S}$.
\Until CONVERGE
\end{algorithmic}
\end{algorithm}

\begin{figure*}
    \centering
    \includegraphics[width=\linewidth]{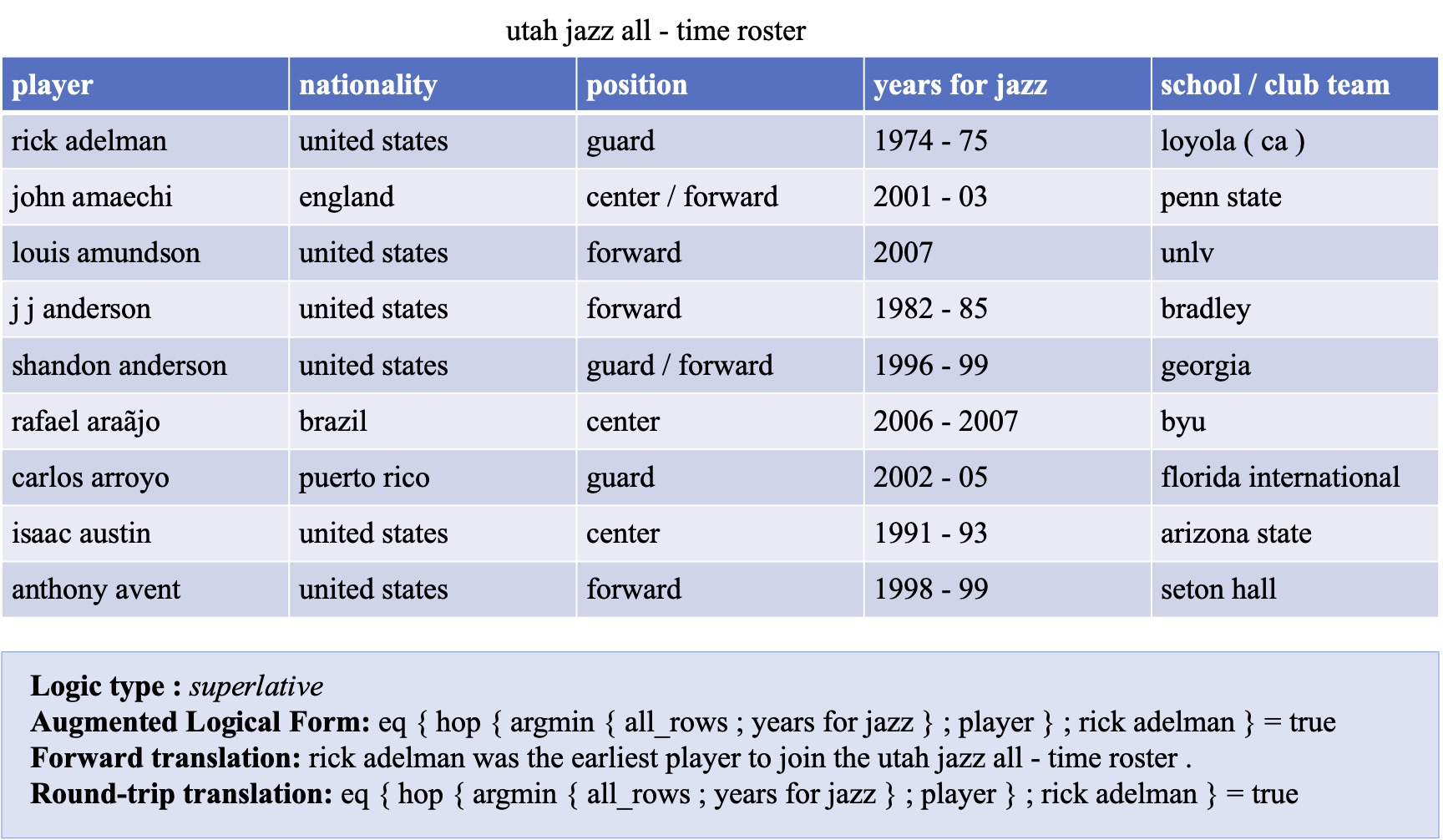}
    \caption{An example of an augmented logical form generated via TopicDA.}
    \label{fig:aug-eg1}
\end{figure*}

\begin{figure*}
    \centering
    \includegraphics[width=\linewidth]{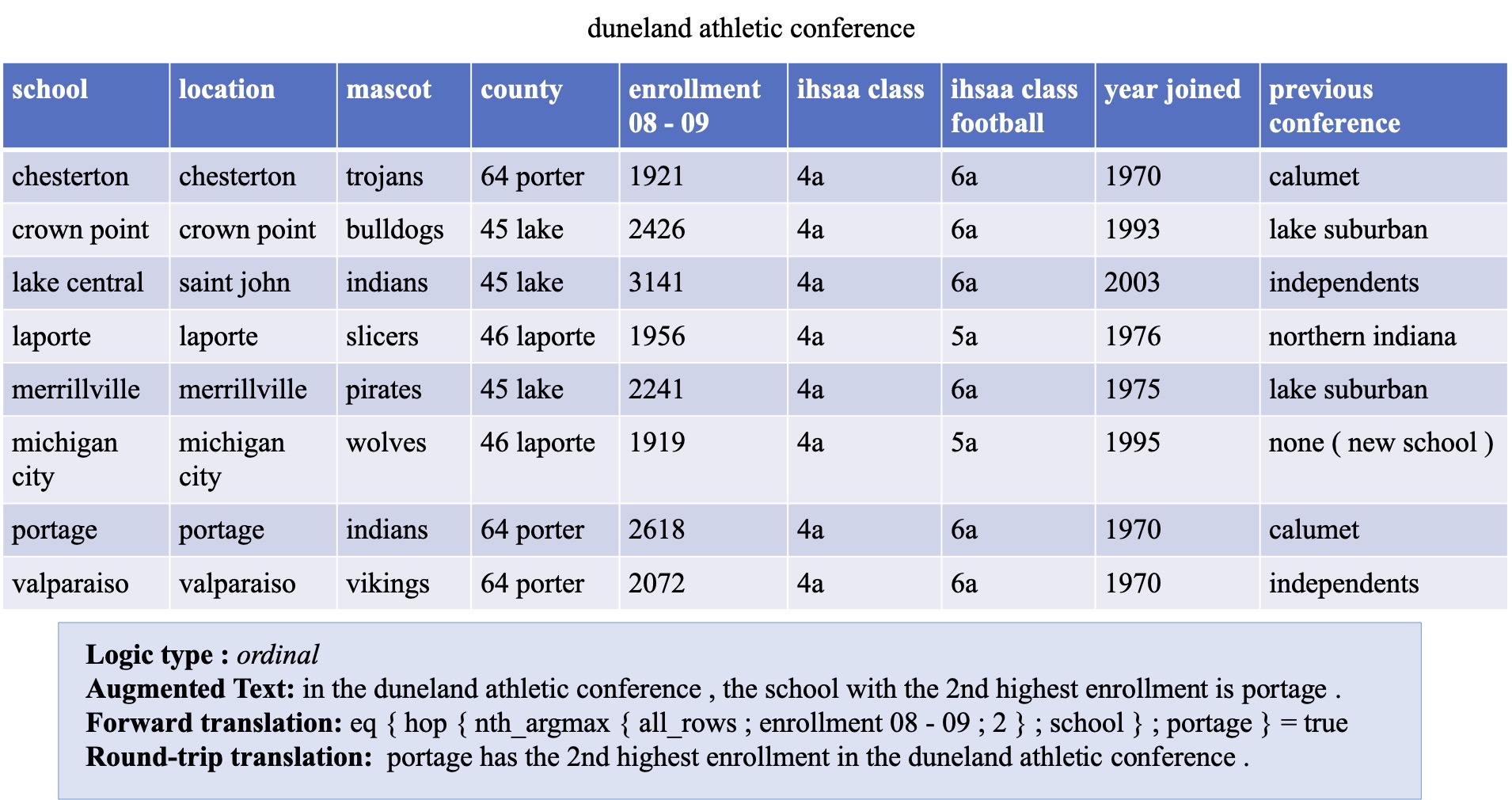}
    \caption{An example of an augmented text generated via TopicDA.}
    \label{fig:aug-eg2}
\end{figure*}

\begin{figure*}
    \centering
    \includegraphics[width=\linewidth]{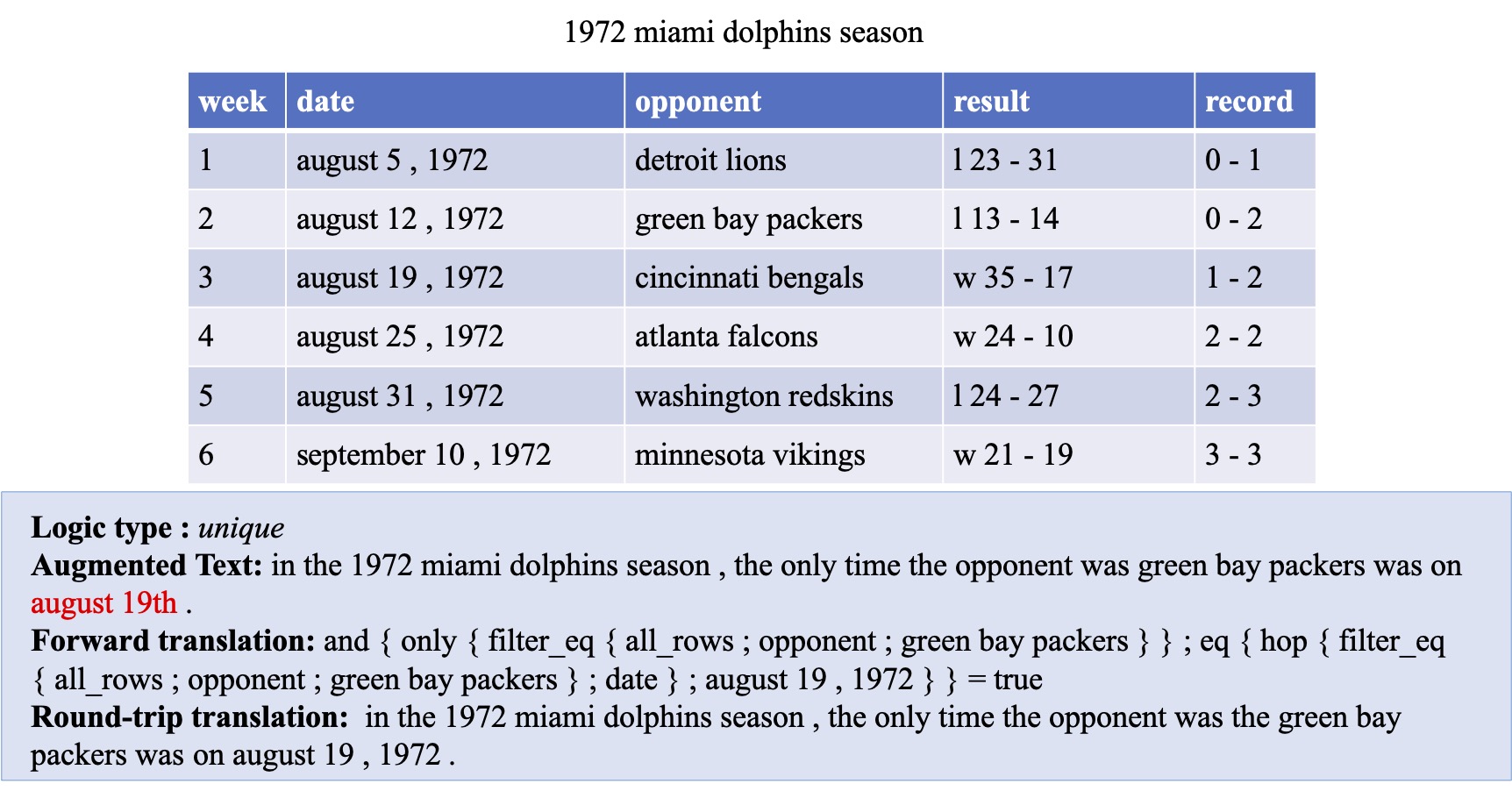}
    \caption{An incorrect example of an augmented text generated via TopicDA.}
    \label{fig:aug-eg3}
\end{figure*}

\begin{table}[t]
    \centering
    \resizebox{\columnwidth}{!}{
    \begin{tabular}{lcc}
    
    \toprule
      Models   & IAA w.r.t. Factual ratings. & IAA w.r.t. Semantical ratings \\
      \midrule
     Gold    &  0.15 & 0.28 \\
    GPT-2 (full) & 0.52 & 0.47 \\
    Ours (full) & 0.63 & 0.58 \\
    GPT-2 (1k) & 0.64 & 0.66 \\
    Ours (1k) & 0.66 & 0.68 \\
    Avg. & 0.52 & 0.53 \\
    
    \bottomrule
    \end{tabular}
    }
    \caption{IAA w.r.t. the ratings of each entry in Table \ref{tab:humaneval}}
    \label{tab:iaa}
\end{table}

\section{Qualitative Examples of TopicDA}
\label{sec:aug-eg}
We list two examples of our TopicDA method, which show perfect quality in terms of topic consistency and logical fidelity, i.e., factually supported by the tables. In Figure \ref{fig:aug-eg1}, we show the augmented logical form generated by assigning logic type \textit{superlative} to the given table, which describes correct information in the table and matches the given topic. We also list its forward and round-trip translations here, which were used to compute its round-trip data weight. These translations are of considerable quality, indicating a high data weight during semi-supervised training. Similarly, Figure \ref{fig:aug-eg2} demonstrates a good example of text augmentation and its translations. However, sometimes the DA models can generate factually incorrect outputs. As shown in Figure \ref{fig:aug-eg3}, although the augmented text is consistent with the assigned topic ``unique'', the correct date should be ``august 12'' instead of ``august 19''. Nonetheless, its forward translation logical form is syntactically correct and semantically consistent with the text, which suggests that this augmented text may still benefit the models on the translation between logcial forms and texts. Our data weighting method will also assign a high score to this example because the round-trip translation perfectly matches the original text.

\begin{figure*}[t]
    \centering
    \includegraphics[width=\linewidth]{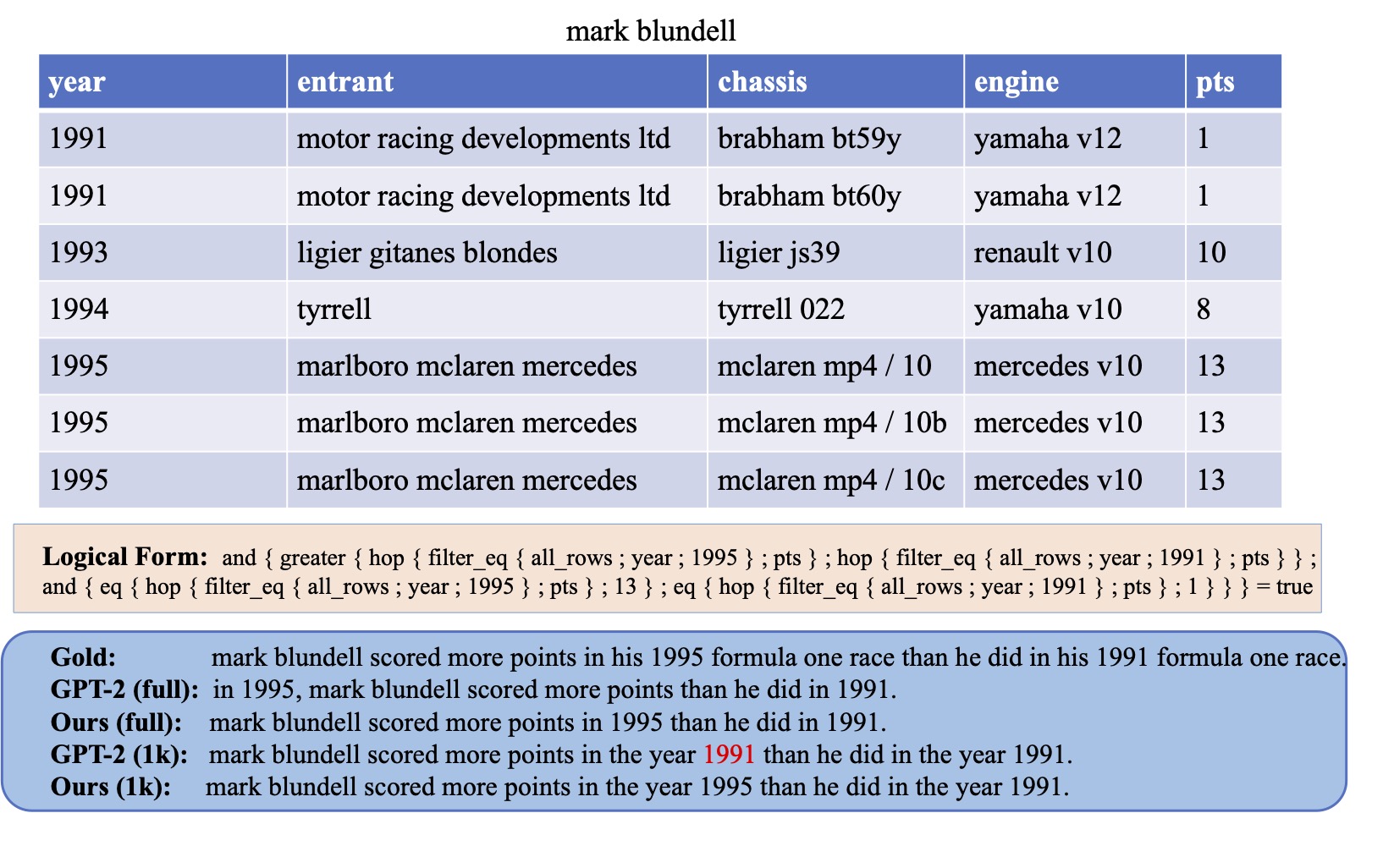}
    \caption{Example one of Logic2text. The incorrect contents are marked as red.}
    \label{fig:case1}
\end{figure*}

\begin{figure*}[t]
    \centering
    \includegraphics[width=\linewidth]{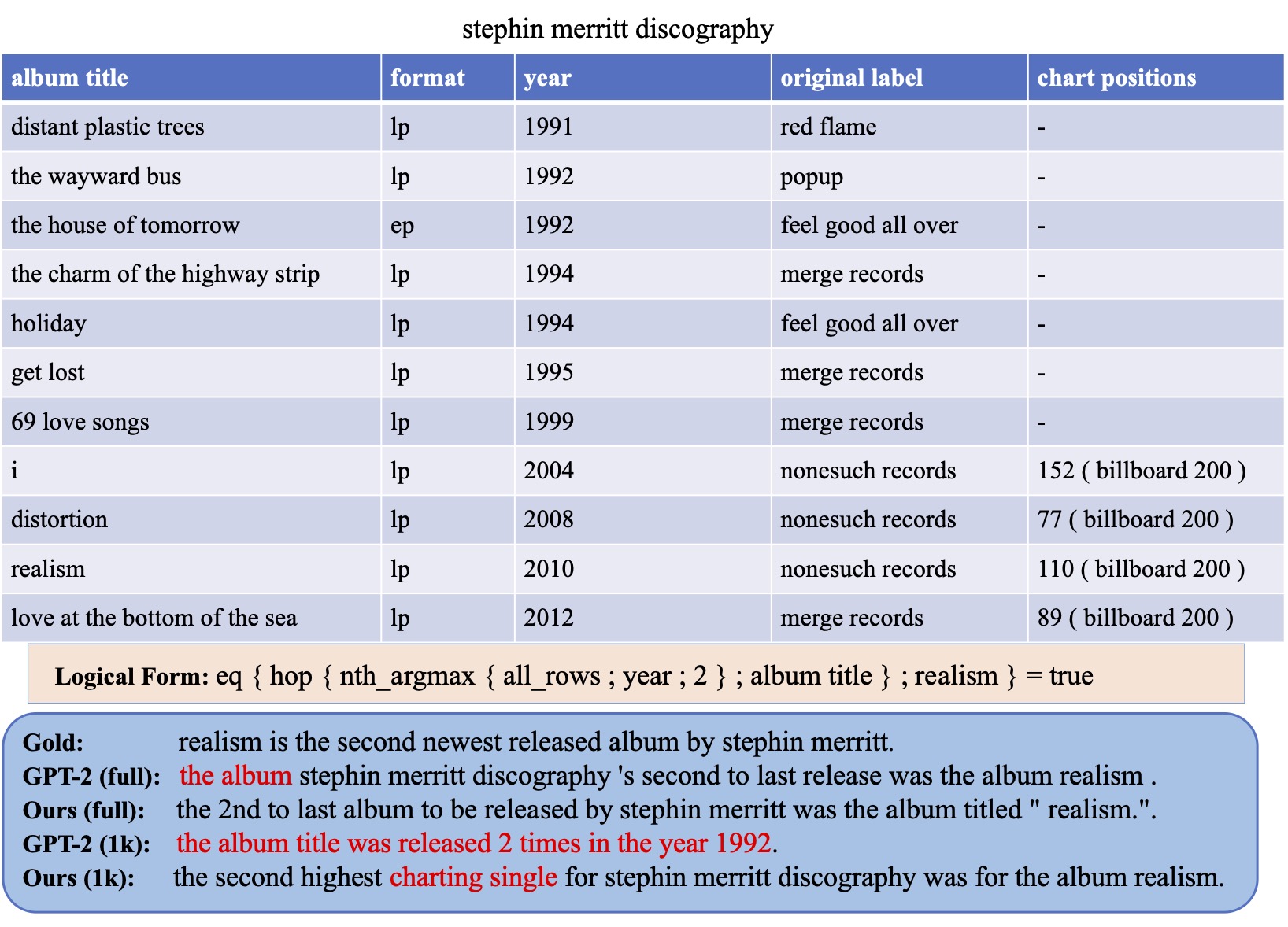}
    \caption{Example two of Logic2text. The incorrect contents are marked as red.}
    \label{fig:case2}
\end{figure*}

\section{Qualitative Examples of Logic2text}
Here, we demonstrate some qualitative examples generated from random samples in the test set of Logic2text. We list the generated descriptions of the five models we compared in human evaluation: Gold, GPT-2(full), Ours(full), GPT-2(1k) and Ours(1k). In Figure \ref{fig:case1}, we observe exciting fluency and fidelity of most generations, except the generation of GPT-2(1k) that mistakenly replaces ``1995'' with ``1991'', possibly because GPT-2(1k) was confused by the multiple entities named ``1991'' appearing in the complicated logical form. In Figure \ref{fig:case2}, the logical form belongs to the \textit{ordinal} logic type, which is more difficult for a model to identify its logical semantics. As a result, GPT-2 (full) failed to generate a factually correct sentence because it mistakenly treated ``stephin merritt discography'' as an album. In contrast, the generation of Ours(full) is both factually and semantically correct. Ours (1k) failed to detect the key column name ``year'' and generated hallucinated words ``charting single''. However, it still interpreted the meaning of function ``nth\_argmax'' in the logical form as ``second highest'', while GPT-2 (1k) totally misunderstood the meaning of the logical form and generated a hallucinated sentence. We can observe that although these GPT-2 based models can basically generate highly fluent sentences, they are not robust enough to generate descriptions faithful to both the table and the logical form. However, these examples demonstrate the effectiveness of our data augmentation framework and using more training examples. We can also find that the logical forms make it easier to interpret the errors made by the models than direct table-to-text generation. It is also interesting how the models translate the compositional logical forms into natural language, and it could be a promising direction to study the compositional generalization problem of Logic2text.

\section{Details of Human Evaluation}
In Table \ref{tab:iaa}, we report the Inter-Annotator Agreement (IAA) scores (Fleiss' Kappa) among our three annotators \textit{w.r.t. their ratings of each entry in Table 4}. The averaged IAA scores are 0.52 w.r.t. the rating of factual correctness and 0.53 w.r.t. semantical correctness, showing a reasonable agreement. The IAA scores are much lower for the rating of Gold References, probably because the errors in gold references are rare and subtler to judge. 


\end{document}